\documentclass[conference,A4]{IEEEtran}
\usepackage{pifont}% http://ctan.org/pkg/pifont
\IEEEoverridecommandlockouts
\usepackage{cite}
\usepackage{amsmath,amssymb,amsfonts}
\usepackage{algorithmic}
\usepackage{graphicx}
\usepackage{textcomp}
\usepackage{xcolor}
\usepackage{multirow}
\usepackage{hyperref}
\usepackage{booktabs}
\usepackage{siunitx}
\usepackage{tikz}
\usetikzlibrary{arrows.meta, shapes, positioning}
\def\BibTeX{{\rm B\kern-.05em{\sc i\kern-.025em b}\kern-.08em
    T\kern-.1667em\lower.7ex\hbox{E}\kern-.125emX}}
    
\begin{document}

\makeatletter % changes the catcode of @ to 11
\newcommand{\linebreakand}{%
  \end{@IEEEauthorhalign}
  \hfill\mbox{}\par
  \mbox{}\hfill\begin{@IEEEauthorhalign}
}
\makeatother

\title{Scene Understanding in Pick-and-Place Tasks: Analyzing Transformations Between Initial and Final Scenes}

\author{\IEEEauthorblockN{Seraj Ghasemi\IEEEauthorrefmark{1},
Hamed Hosseini\IEEEauthorrefmark{1},
MohammadHossein Koosheshi\IEEEauthorrefmark{1},
Mehdi Tale Masouleh\IEEEauthorrefmark{1}, and
Ahmad Kalhor\IEEEauthorrefmark{1}}
\IEEEauthorblockA{\IEEEauthorrefmark{1}Human and Robot Interaction  Lab, School of Electrical and Computer Engineering,  \\ 
University of Tehran, Tehran, Iran}}

% \author{
% \IEEEauthorblockN{Seraj Ghasemi}
% \IEEEauthorblockA{\textit{Human and Robot} \\ \textit{Interaction Laboratory} \\
% \textit{School of Electrical and }\\
% \textit{Computer Engineering}\\
% \textit{University of Tehran}\\
% seraj.ghasemi@ut.ac.ir}
% \and
% \IEEEauthorblockN{Hamed Hosseini}
% \IEEEauthorblockA{\textit{Human and Robot} \\ \textit{Interaction Laboratory} \\
% \textit{School of Electrical and }\\
% \textit{Computer Engineering}\\
% \textit{University of Tehran}\\
% hosseini.hamed@ut.ac.ir}\and 
% \IEEEauthorblockN{MohammadHossein Koosheshi}
% \IEEEauthorblockA{\textit{Human and Robot} \\ \textit{Interaction Laboratory} \\
% \textit{School of Mechanical}\\
% \textit{ Engineering}\\
% \textit{University of Tehran}\\
% mhkoosheshi@ut.ac.ir}\and
% \linebreakand
% \IEEEauthorblockN{Mehdi Tale Masouleh}
% \IEEEauthorblockA{\textit{Human and Robot} \\ \textit{Interaction Laboratory} \\
% \textit{School of Electrical and }\\
% \textit{Computer Engineering}\\
% \textit{University of Tehran}\\
% m.t.masouleh@ut.ac.ir}\and
% \IEEEauthorblockN{Ahmad Kalhor}
% \IEEEauthorblockA{\textit{Human and Robot} \\ \textit{Interaction Laboratory} \\
% \textit{School of Electrical and }\\
% \textit{Computer Engineering}\\
% \textit{University of Tehran}\\
% akalhor@ut.ac.ir}}

\maketitle

\begin{abstract}
With robots increasingly collaborating with humans in everyday tasks, it is important to take steps toward robotic systems capable of understanding the environment. This work focuses on scene understanding to detect pick-and-place tasks given initial and final images from the scene. To this end, a dataset is collected for object detection and pick-and-place task detection. A YOLOv5 network is subsequently trained to detect the objects in the initial and final scenes. Given the detected objects and their bounding boxes, two methods are proposed to detect the pick-and-place tasks which transform the initial scene into the final scene. A geometric method is proposed which tracks objects' movements in the two scenes and works based on the intersection of the bounding boxes which moved within scenes. Contrarily, the CNN-based method utilizes a Convolutional Neural Network to classify objects with intersected bounding boxes into 5 classes, showing the spatial relationship between the involved objects. The performed pick-and-place tasks are then derived from analyzing the experiments with both scenes. Results show that CNN-based method, by VGG-16 backbone, outscores the geometric method by roughly 12\% in certain scenarios, with an overall success rate of 84.3\%. 
\end{abstract}


\section{Introduction}
Humans can effortlessly discern pick-and-place operations executed between two initial and final frames of a scene. In order to enable a robot of the same, the rearrangement challenge of AI2-THOR \cite{kolve2017ai2} and CVPR was introduced in 2022 to the end of rearranging a scene to match a goal state and is still an open problem in the field of robotics and manipulation \cite{ai2thor}. Moreover, with the advancements in deep learning algorithms, robotic manipulation especially robotic grasping entered a new era. A successful robotic grasp candidate strongly depends on the task the robot intends to perform \cite{balazade}\cite{anushe}. As a result, task identification takes place before task execution. Tasks are usually identified in a predefined way, but in a more realistic approach simillar to packaging a quantity of scattered chocolates \cite{mojtahedi2024experimental}, the robotic task can be derived from the environment that a robot is exposed to. In other words, if a robot sees an arranged scene and is then exposed to an unarranged one, it should be able to identify the required robotic tasks which can transform the unarranged scene into an arranged one. This work focuses on training a robotic manipulator to process the arranged and unarranged frames related to an environment to produce an optimized pick-and-place strategy. In order to complete the pick-and-place operation, the object needs to be grasped by a grasp detection approach which is widely discussed in literature \cite{hosseini2020improving}\cite{agile}. The proposed method hinges on a scene understanding algorithm based on object detection which is followed by pick-and-place operation. Therefore, in the following paragraphs, the relevant literature on scene understanding, and object detection in robotic operation is reviewed to provide a framework for the upcoming discussion.

\begin{figure}[!tp]
\centerline{\includegraphics[ width = 0.35\textwidth]{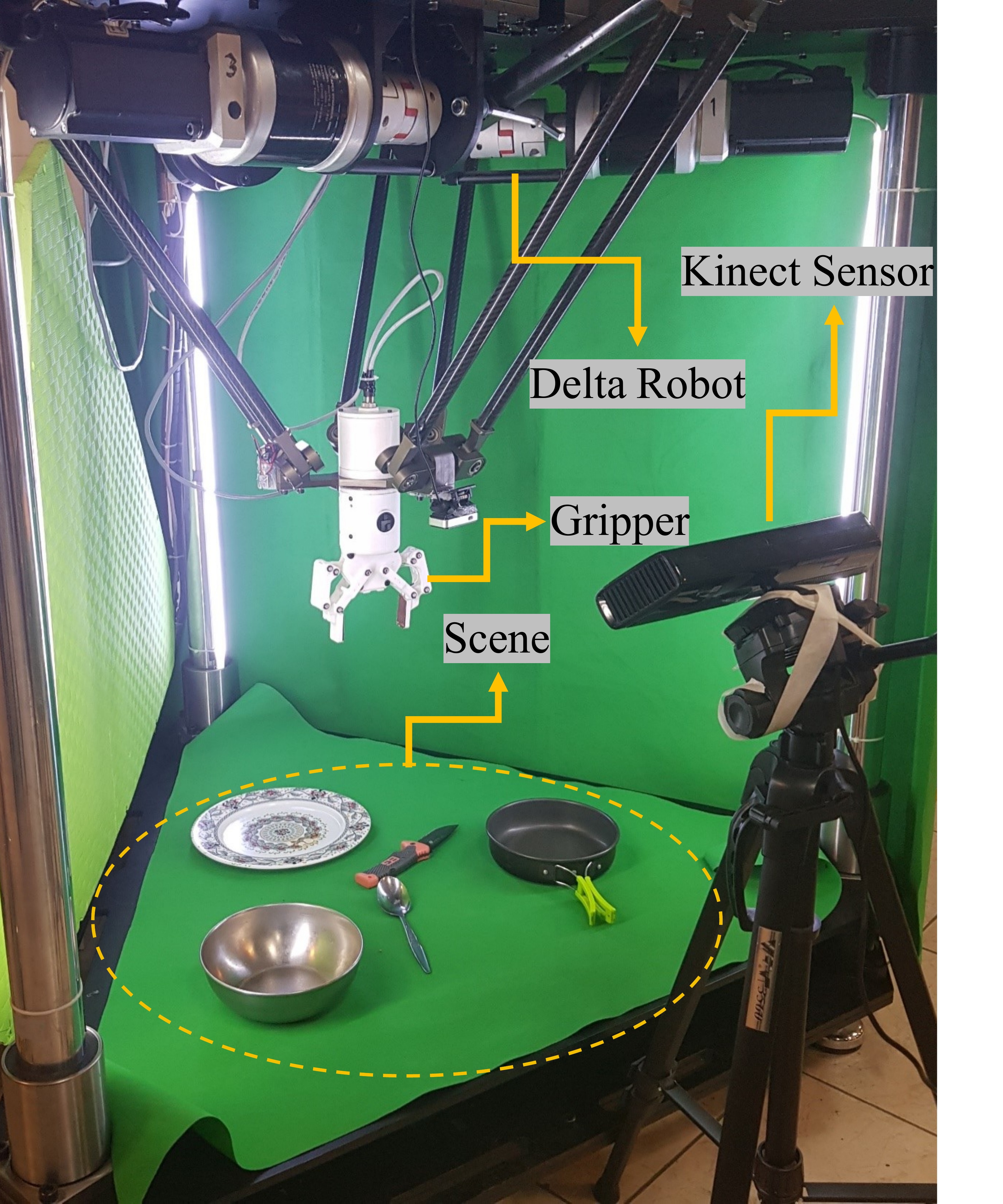}} 
\caption{The pick-and-place setup used in this study and its components.}
\label{fig:setup}
\end{figure}

\begin{figure*}[!tp]
\centerline{\includegraphics[ width =0.7\textwidth]{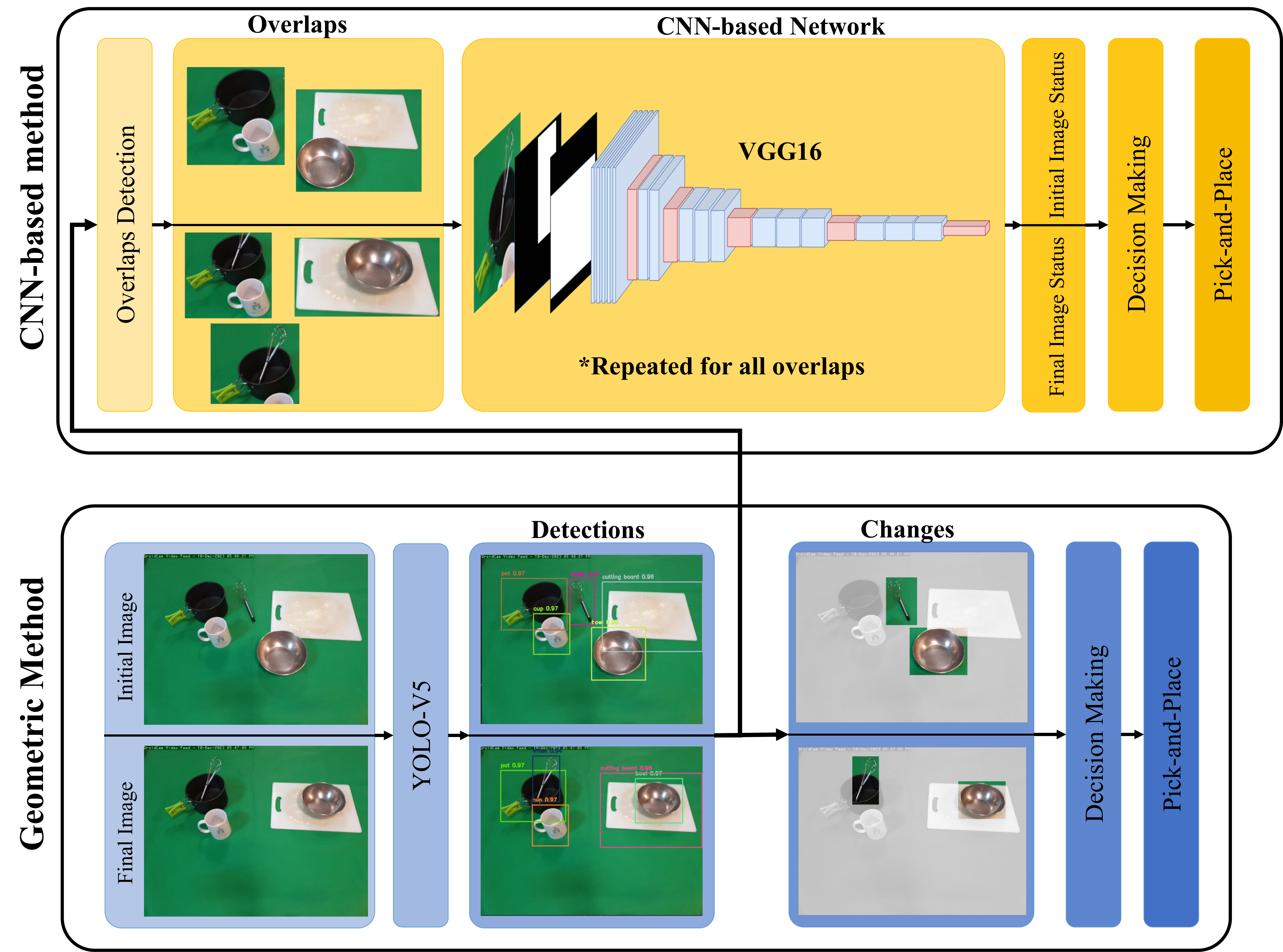}}
\caption{Overall flow of the proposed methods. Both geometric and CNN-based methods use object detection results. The geometric method tracks the bounding boxes in the initial and final images, while the CNN-based method uses a CNN network to detect spatial relationships between objects whose bounding boxes overlap and identify the pick-and-place operations that transform the initial image into the final image.}
\label{fig:graphical_abstract}
\end{figure*}

\textbf{Scene Understanding}.
Robots need to recognize object affordances to understand and interact with the objects in their environment \cite{zech2017computational}. The majority of earlier affordance detection research has employed RGB-D pictures or point cloud data for grasp detection \cite{bohg2009grasping}. These techniques can result in grasping effective motions, but they fall short of providing the robot with additional visual information for handling the object in a human-like manner \cite{2024Naeinian}. Also, the previous models have explored improving robots' understanding of spatial relationships between objects within an image, leveraging Graph Convolutional Networks(GCN) \cite{wu2023prioritized}. In contrast to an object's visual or physical attributes, which describe the property of the object, affordances show how components of an object functionally interact with a robot for a specific task. In reality, the essential information which is required for manipulation tasks is derived from object affordance. For instance, when the robot is supposed to transfer water from a bottle into a bowl, it should be able to both identify and locate the appropriate components (bottle, bowl) as well as their affordances (grab, contain). In the present work, object affordances from two pictures are examined at the pixel level. Consequently, the work of identifying object affordance may be viewed as a continuation of the widely recognized computer vision problem of semantic object detection. Nevertheless, since object affordances convey the abstract of how people interact with objects, it poses a more challenging issue than the standard detection problem. Knowing object affordances empowers the robot to make more independent decisions about the course of action to take for each manipulation task \cite{zech2017computational}. In \cite{nguyen2019scene}, the affordance detection issue is equivalent to a visual video translation task where the objective is to convert a given video into a command, mainly inspired by video captioning topics \cite{donahue2015long} \cite{venugopalan2015sequence}.

\textbf{Object Detection.}
Different object detection algorithms are used in the field of robotics and one of the most popular ones is YOLO networks whose evolution path is explained by \cite{terven2304comprehensive}. YOLO-based object detection finds applications in diverse robotic tasks like quadcopter planning \cite{rahmania2022exploration}, autonomous underwater vehicles navigation \cite{zhao2022improved}, controlling a mobile robot \cite{li2023yolo} and employing service robots\cite{ye2023dynamic}. Some works modify the YOLO network architecture in order to adapt it to their desired applications \cite{ahmad2020object} \cite{kshirsagar2023modified}. Some others are based on using predicted labels and bounding boxes of the network directly in their robotic task \cite{aljaafreh2023real}. The current paper uses the YOLO network as an intermediary stage in the inference flow, as the output classes and bounding boxes of the YOLO network are fed into the subsequent networks and stages in the scene understanding flow.

The main contribution of this paper is providing a framework for autonomous robotic manipulation without explicitly requiring human command and by providing the final desired scene for the robot. In other words, this work proposes a method to analyze the current and the final scenes to derive the required pick-and-place tasks for transforming the initial scene into the final scene. To this end, two methods are proposed which are built on object detection. Given the detected objects in the scene, a geometric method is proposed to process the movement and the overlap of the bounding boxes to detect the performed pick-and-place tasks. Alternatively, a CNN-based method is suggested which utilizes a CNN capable of understanding the spatial relationship between the objects whose bounding boxes have an intersection, which are possible candidates for a performed pick-and-place task. Finally, the pick-and-place tasks are performed using a grasp detection method based on primitive shape segmentation \cite{agile}\cite{hamed_primitive} in a practical setup which is depicted in Fig. \ref{fig:setup}. The inference flow of the proposed methods is illustrated in Fig. \ref{fig:graphical_abstract}. As illustrated in Fig. \ref{fig:graphical_abstract}, both methods make use of the object detection results. The geometric method uses analytical methods to predict the pick-and-place tasks, and the CNN-based method uses a CNN to detect the spatial relationship between objects in each image \cite{zhu2018learning} and subsequently detects the performed pick-and-place operations.

This paper is organized as follows. First, the required dataset and its collection and labeling processes are discussed. Second, both geometric and CNN-based pick-and-place task detection methods, the grasping methods, and the practical setup are presented. Lastly, the results for both methods are provided and compared to conclude the paper.

% related work
% introduction

% CNN
% CNN figures
% CNN labels
% balancing
% train balancing between classes with aug
% grasping methods

% results
% learning curve (*)
% confusion (*)
% success and unsuccess cases for both methods.
% compare accuracy for both methods
% plot grasps

\begin{figure}[!tp]
\centerline{\includegraphics[ width = 0.5\textwidth]{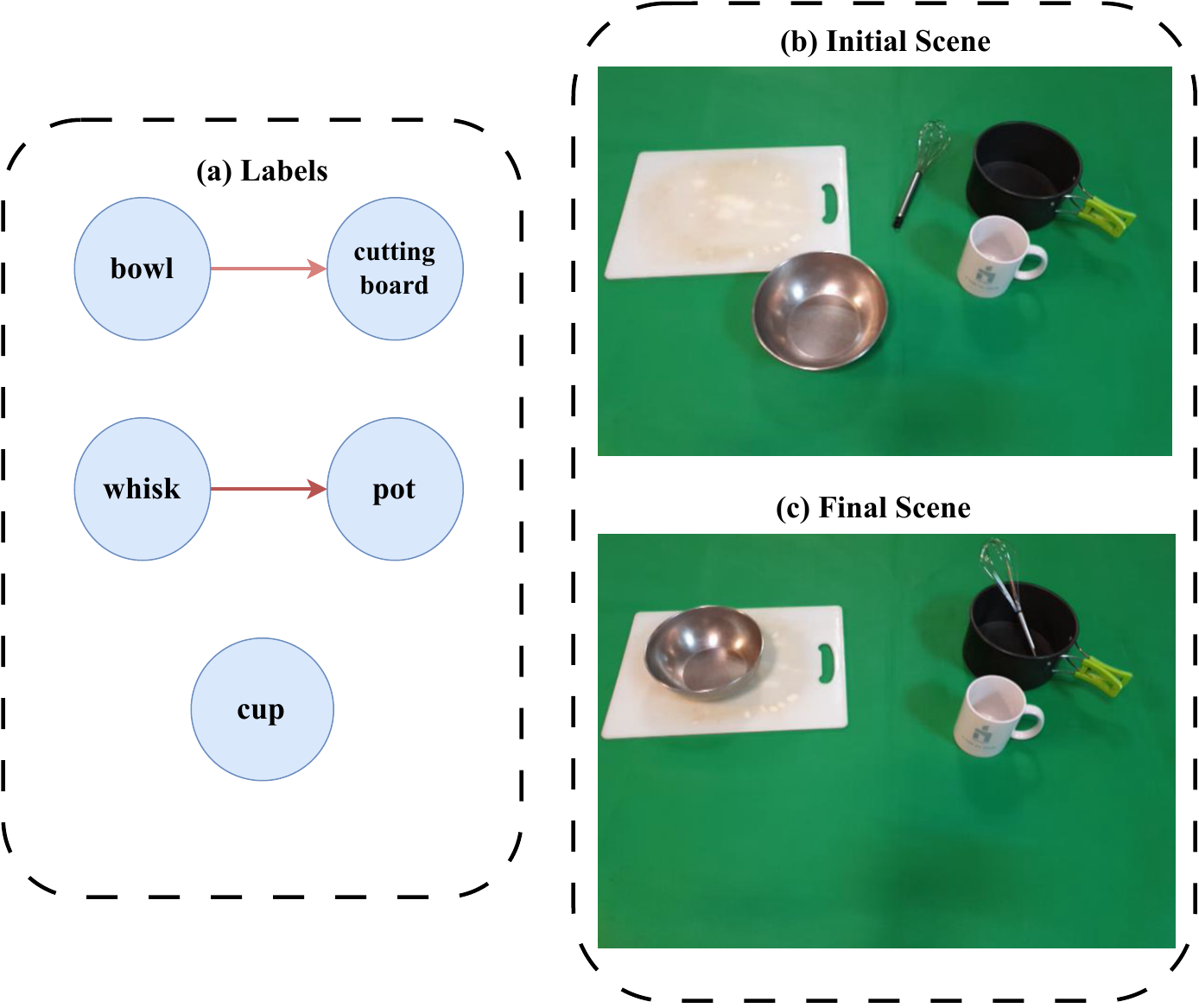}}
\caption{Dataset collection procedure. In (a) a set of objects and pick-and-place tasks are presented by the GUI, (b) shows the initial scene image set up by the user, and (c) shows the scene after the performed pick-and-place tasks.}
\label{fig:datagraph}
\end{figure}

\section{Datasets and Object Detection}
\label{sec:objdet}

In order to train the object detection network, a dataset of 11 objects is collected. The objects included in this dataset are limited to household objects used in kitchens. Each dataset input consists of two images, the initial and the final scenes. For each input, several pick-and-place tasks are performed to transform the initial image into the final.

A Graphical User Interface (GUI) is designed for data collection. The GUI randomly picks a few objects and defines pick-and-place tasks between them. Subsequently, the user sets the scene, in which the selected objects by the GUI are available in the first image with random placements. Then, the pick-and-place tasks are performed and the final image is set and recorded. Fig. \ref{fig:datagraph} shows a pair of input and output images along with the corresponding selected pick-and-places by the GUI. Following this procedure, 224 images are collected, 112 as initial and final images. Each image is also annotated for the object detection task using the Roboflow annotation tool \cite{Roboflow}. As a result, 967 bounding boxes are annotated in the 224 images.

In order to detect objects in the scene, YOLOv5 \cite{redmon2016you} is trained using the collected dataset. The architecture of YOLOv5 is characterized by its use of the CSP (Cross Stage Partial) network as the backbone, which helps in reducing the computational cost while maintaining efficiency. YOLOv5 also employs the PANet (Path Aggregation Network) for feature aggregation, enhancing its ability to detect objects at different scales. Additionally, YOLOv5 incorporates multiple head layers responsible for predicting objects of various sizes. This streamlined design allows YOLOv5 to achieve high frame rates in real-time applications, making it suitable for scenarios that require rapid and accurate object detection. It is proven to be robust, accurate, and efficient, YOLOv5 stands out as one of the most effective models for object detection tasks. As a result, this work employs the fifth version of YOLO as its object detection method. The YOLOv5 network is trained for 250 epochs using Adam optimizer and an initial learning rate of 0.01. In order to augment the 224 images in the training set, various image transformations are applied to triple the set. These transformations include rotation, translation, scaling, mosaic, mixup, HSV shift, and horizontal flip.

\section{Pick-and-Place Task Detection}
This section discusses two methods to detect the performed pick-and-place operations, namely the geometric and CNN-based methods. Starting from the initial and final images, both methods utilize bounding boxes predicted by an object detection model, trained on the collected dataset presented in Section \ref{sec:objdet}. The geometrical method is based on a non-learning approach to detect which pick-and-place tasks transform the initial scene into the final. Contrarily, the CNN-based method uses a learning approach. Subsequently in this section, both methods are explained in detail.

\subsection{Geometric Pick-and-Place Task Detection}
\label{sec:geomethod}
A straightforward approach to detect the performed pick-and-place tasks is to track their bounding boxes and employ geometrical methods to analyze the movements leading to pick-and-place tasks. A procedure is proposed to track objects and detect the pick-and-place tasks performed in the scene by analyzing the initial and final images from the scene. This process begins by both initial and final images are fed into the trained YOLOv5 network, which yields the bounding boxes of objects present in the scene. For each object in the scene, the movement of their bounding box is calculated. To associate the movement with a performed pick-and-place task, and to reduce the chances of detecting minor movements and a shake of the camera as a pick-and-place task, a threshold is set to obtain the objects that are moved. Then, the Intersection Over Union (IOU) of the moved objects' bounding boxes with those of other objects is calculated to derive the next object involved in the pick-and-place task. Notably, there is another threshold of 20\% set for the calculated IOUs to detect the placement of an object on another. Given the pairs of objects which participated in a pick-and-place task, the object in each pair with the more substantial movement is considered the picked object, placed on/in the other object.

\subsection{CNN-based Pick-and-Place task Detection}
\label{sec:cnnmethod}
A CNN is trained to analyze the scene. Given the bounding boxes of each pair of objects in the scene generated by YOLOv5, the classification model based on CNN processes the information to detect the spatial relationship statuses that can happen as a result of a pick-and-place task.  

As aforementioned, the CNN-based method makes use of the bounding boxes given by the YOLOv5 network. In order to feed the network during training and inference, the bounding boxes of each pair of objects that have non-zero IOU are used, since the IOU value of more than zero between these classes can be a sign of having a spatial relationship. The smallest bounding box which contains both objects is used to crop the input RGB image. The cropped RGB image is then concatenated with two binary masks to make up inputs of channel size 5. Each pixel value of the binary mask is 1 if the pixel is in the object's bounding box. The first three channels of the concatenated data are the cropped image and the next channels are the binary masks of the first and second objects respectively.

Since a pick-and-place task can result in an object being "in" or "on" another object, it is important to separate these two cases. Furthermore, for each "on" and "in" class, there are two classes indicating which object is picked and placed on or in the other object, which results in a total of 4 classes. Another class showing simply no spatial relationship information between objects is added to the 4 mentioned classes. The CNN utilizes backbones pre-trained on the imagenet dataset \cite{deng2009imagenet}, a subsequent flattening layer, and a dense layer with a size of 5 as the classifier. Three backbones are tested for the CNN, which are Resnet-50 \cite{he2016deep}, Resnet-101 \cite{he2016deep}, and VGG16 \cite{simonyan2014very} networks.

As discussed, the CNN is responsible for detecting the spatial relationship between two objects in a scene that have bounding boxes with IOU more than zero. In order to detect the performed pick-and-place operations, the initial and the final scenes are fed into the CNN separately. If the outputs of the two experiments with the CNN result in two different classes, a pick-and-place is assigned to the two bounding boxes. For example, if the first experiment with CNN results in the class "Mask 1 in Mask 2" and the second experiment bears "Unrelated", it is inferred that the object of the first mask is picked and placed out of the object of the second mask.

\subsection{Grasping methods and setup}
The practical setup to perform pick-and-place is depicted in Fig. \ref{fig:setup}. The gripper used in this study is a two-fingered robotic gripper fabricated by \cite{navidhand2023}. Also, the Delta parallel Robot is one of the most popular manipulators in pick-and-place robotic tasks, and this study utilizes a 3-DOF Delta parallel Robot \cite{tamizi2022experimental}. A Kinect v1 sensor is included in the setup as the camera in this study, as it is capable of recording images in RGB and Depth modalities.

Given the detected pick-and-place tasks, and starting from the initial scene, it is important to grasp accordingly. For grasp detection purposes, the methods discussed in \cite{hamed_primitive} are employed. Accordingly, each object is segmented into its primitive shapes, and the point cloud of the primitive shapes is generated. Subsequently, utilizing reference primitive shape point clouds and their predefined grasping candidates, the suitable grasp is detected and performed. After grasping, the picked object is moved to the center of the bounding box and the object is released to be placed in the destination. Notably, this study follows the calibration procedure proposed in \cite{hamed_primitive} using point clouds and for both grasping and placing stages.

\section{Results}
In this section, the results of the object detection and the performance of each pick-and-place task detection method are studied. Regarding the pick-and-place task detection methods, the results obtained by three different CNN-based pick-and-place task detection networks are discussed. Then, the best network is compared to the geometric method to study how the CNN can overcome the problems that the geometric method is unable.

\subsection{Object Detection Results}
It is important to study the performance of the YOLOv5 since both pick-and-place task detection methods depend on a robust and accurate object detection performance. Following the training procedure discussed in Section \ref{sec:objdet}, Table \ref{tab:objdet} shows object detection results over unseen test images. According to Table \ref{tab:objdet}, the YOLOv5 model obtains a mean average precision (mAP) of 97.6\% overall.

\begin{table}[!tp]
\centering
\caption{Object detection results for each class.}
\label{tab:objdet}
\begin{tabular}{@{}lcccccc@{}}
\toprule
Class & Instances & Precision & Recall & mAP \\ 
\midrule
Overall   &  190       & 0.965 & 0.996 & 0.976\\
Bottle     & 6         & 0.851 & 1     & 0.853 \\
Pan   & 14 & 0.996 & 1 & 0.995 \\
Plate & 15 & 0.934 & 1 & 0.995 \\
Pot   & 9  & 0.893 & 1 & 0.94 \\
Spoon & 19 & 0.995 & 1 & 0.995 \\
Whisk & 17 & 0.962 & 0.941 & 0.947 \\
Knife & 20 & 0.997 & 1 & 0.995 \\
Bowl  & 42 & 0.975 & 0.952 & 0.954 \\
Cup   & 30 & 0.926 & 0.9   & 0.902 \\
Cutting board & 48 & 0.997 & 1 & 0.995 \\
Fork & 35 & 0.916 & 0.829 & 0.896 \\
\bottomrule
\end{tabular}

\label{your-label-here}
\end{table}

\subsection{Pick-and-Place Task Detection Results}
\begin{figure*}[!tp]
\centerline{\includegraphics[width=\textwidth]{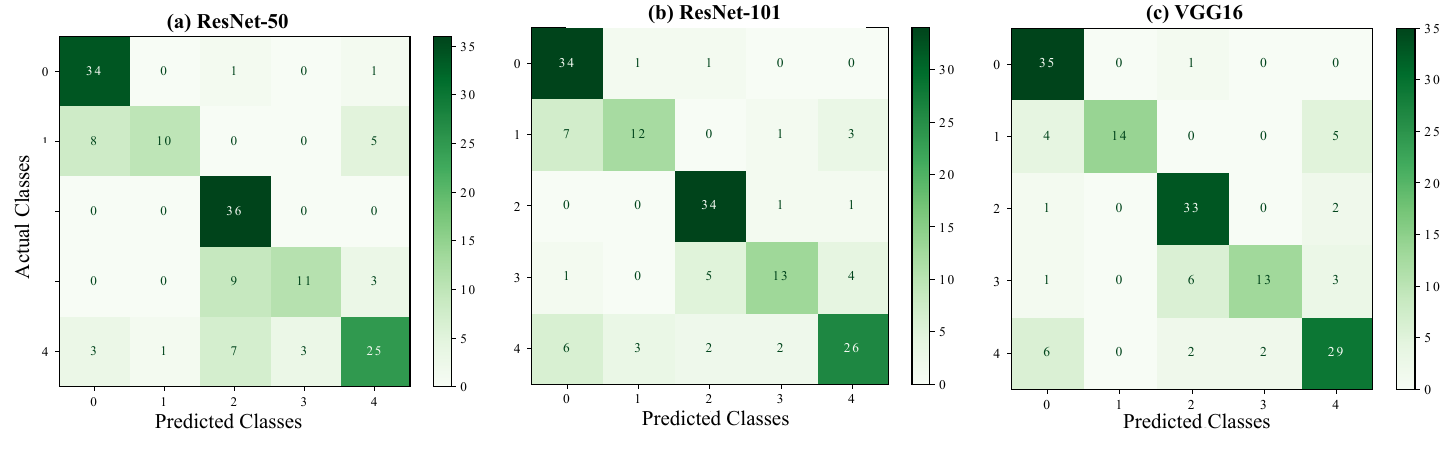}}
\caption{Confusion matrix of classification results over validation dataset for (a) ResNet-50, (b) ResNet-101, and (c) VGG16.}
\label{fig:confusions}
\end{figure*}

\begin{figure*}[!tp]
\definecolor{geo}{RGB}{0,204,0}
\definecolor{cn}{RGB}{204,0,204}
\definecolor{gt}{RGB}{0,0,0}
\centerline{\includegraphics[width=0.8\textwidth]{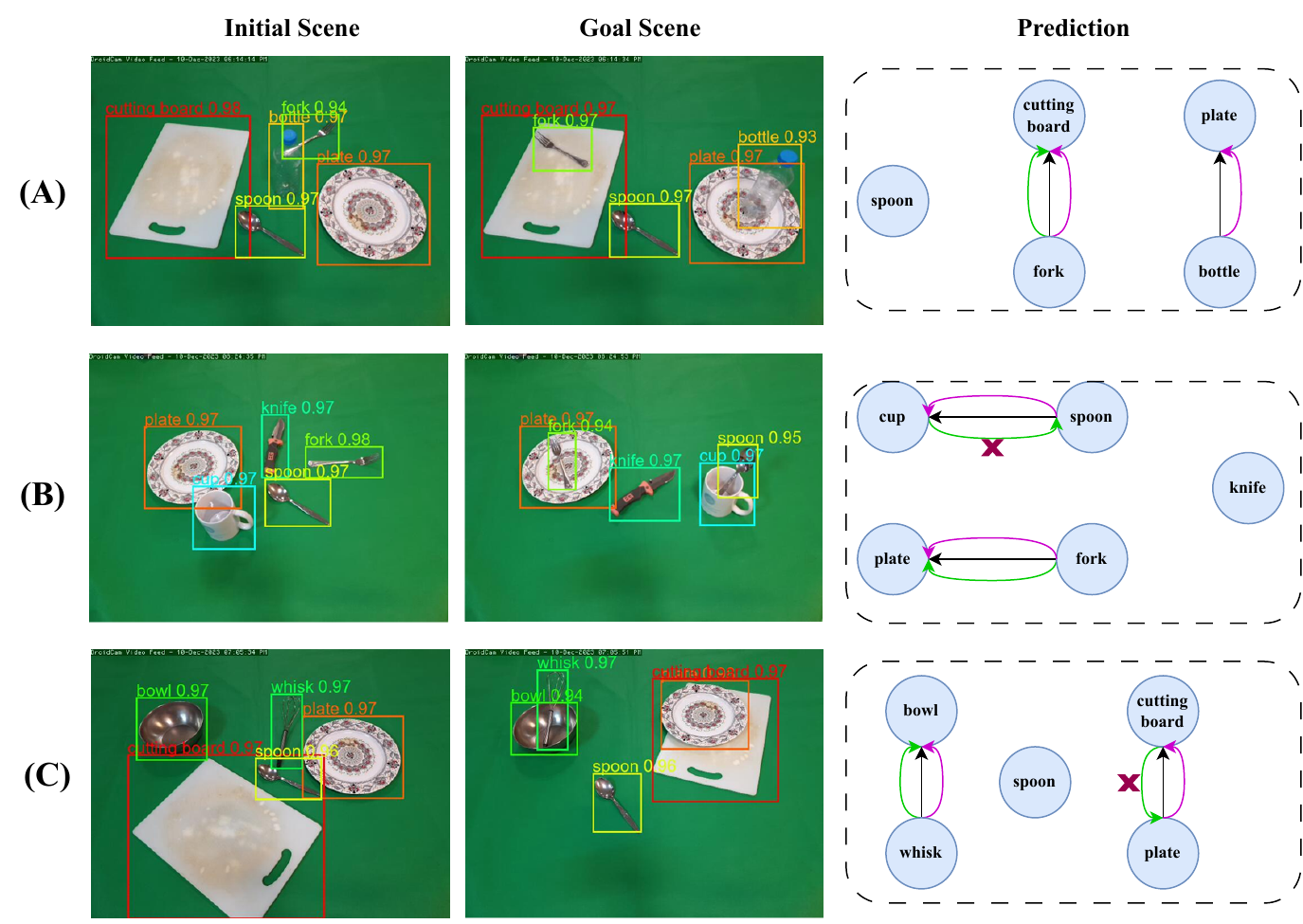}}
\caption{Three unseen initial and final scenes, object detection results and confidence score, and predicted pick-and-place tasks by geometric method displayed with (green arrow ), CNN-based methods displayed with (purple arrow), and the ground truth displayed with (black arrow).}
\label{fig:tests}

% Insert TikZ arrows outside the caption to avoid LaTeX errors
% \centering
% \begin{tikzpicture}
%   \draw[->, geo, line width=1.2pt] (0,0) -- (0.35,0); % Green arrow for geometric method
%   \node at (0.6,0) {(green arrow)};
% \end{tikzpicture}
% \quad
% \begin{tikzpicture}
%   \draw[->, cn, line width=1.2pt] (0,0) -- (0.35,0);  % Purple arrow for CNN-based method
%   \node at (0.6,0) {(purple arrow)};
% \end{tikzpicture}
% \quad
% \begin{tikzpicture}
%   \draw[->, gt, line width=1.2pt] (0,0) -- (0.35,0);  % Black arrow for ground truth
%   \node at (0.6,0) {(black arrow)};
% \end{tikzpicture}
\end{figure*}

The CNN-based pick-and-place task detection network discussed in Section \ref{sec:cnnmethod} is trained using different backbones. The networks are trained using an Adadelta optimizer, Cross Entropy as the loss function, and a training batch size of 32. Table \ref{tab:backbones} shows the classification results of the backbones with the best results over the validation dataset. Task-1 refers to understanding the direction of the pick-and-place, and Task-2 is the distinction between "in" and "on" classes as a result of a pick-and-place. Accordingly, VGG16 yields better results with more than 5\% in precision than ResNet-50. Fig. \ref{fig:confusions} shows the confusion matrix of classification results. Evident in Fig. \ref{fig:confusions}, VGG16 performs better than its ResNet counterparts mainly on the 4th class, which is the class indicating the objects are unrelated. Moreover, VGG16 manages better distinction between classes "in" and "on" classes. VGG16 is subsequently determined as the main backbone for the CNN-based method for further discussion due to its better performance.

\begin{table}[!tp]
  \centering
    \caption{The performance of different backbones of the CNN model over unseen data.}
  \label{tab:backbones}
  \begin{tabular}{|c|c|c|c|c|}
    \hline
    Backbone & Task & Accuracy & Precision & Recall \\
    \hline \hline
    \multirow{3}{*}{ResNet-50} & Task-1 & 85.2 & 43.0 & 43.2 \\
    \cline{2-5}
    & Task-2 & 77.8 & 39.5 & 36.0 \\
    \cline{2-5}
    & Overall & 78.0 & 71.4 & 64.4 \\
    \hline \hline
    \multirow{3}{*}{ResNet-101} & Task-1 & 85.2 & \textbf{43.7} & 44.2 \\
    \cline{2-5}
    & Task-2 & 79.8 & 40.2 & 38.7 \\
    \cline{2-5}
    & Overall & 76.1 & 70.2 & 67.7 \\
    \hline \hline
    \multirow{3}{*}{VGG16} & Task-1 & \textbf{85.7}  & 43.4 & \textbf{45.0} \\
    \cline{2-5}
    & Task-2 & \textbf{82.9} & \textbf{43.2} & \textbf{40.6} \\
    \cline{2-5}
    & Overall & \textbf{79.1} &\textbf{ 76.9 }& \textbf{71.2} \\
    \hline
  \end{tabular}
\end{table}

As discussed in Section \ref{sec:objdet}, each pair of initial and final images in the train and test dataset is associated with a label that denotes all the pick-and-place tasks that can transform the initial scene into the final. In addition to analyzing the performance of different networks, it is useful to investigate the geometric and CNN-based methods' success in predicting the pick-and-place tasks over an unseen test set. The CNN-based method and the geometric method are both tested using 32 unseen images. The CNN-based method yields 84.3\% accuracy over these experiments, while the geometric method's accuracy is 72\%. Table \ref{tab:comparison} shows the performance of each method using Precision and Recall as metrics. Class 0 refers to the first object picked and placed on the second object, class 1 refers to the same but in the opposite direction, and class 2 is simply no pick-and-place task detected.

\begin{table}[!tp]
\centering
\caption{Geometric and CNN-based methods performance.}
\begin{tabular}{@{}lSSSSSS@{}}
\toprule
& \multicolumn{2}{c}{CNN-based} & \multicolumn{2}{c}{Geometric} \\
\cmidrule(lr){2-3} \cmidrule(l){4-5}
Class & {Precision} & {Recall} & {Precision} & {Recall} \\
\midrule
0 & 84.37 & 91.52 & 84.75 & 84.75 \\
1 & 86.88 & 89.83 & 84.75 & 84.75 \\
2 & 80.85 & 70.37 & 70.83 & 70.83 \\
\bottomrule
\end{tabular}
\label{tab:comparison}
\end{table}

To qualitatively discuss the results, Fig. \ref{fig:tests} shows three experiments with the predicted pick-and-place tasks by the CNN-based and geometric methods. In this method, the predictions are illustrated by arrows to show the object which is picked and placed on the other object. The experiment depicted in Fig. \ref{fig:tests} (a) shows a successful prediction by the CNN model and an unsuccessful one by the geometric method. The failure of the geometric method is due to the set thresholds discussed in Section \ref{sec:geomethod}, which proves the limitations of the geometric approach. On the other hand, the CNN-based method shows robust results regardless of the bounding boxes' dimensions. An even more interesting experiment is illustrated in Fig. \ref{fig:tests} (b). In this experiment, the geometric method detects a true pick-and-place task between the spoon and the cup but fails to detect the direction of the task correctly. This experiment also points out another advantage of the CNN-based method over the geometric one. The CNN has learned the concepts of a mug, a spoon, and their engagement in a pick-and-place task, and it almost never predicts a task of placing a mug on a spoon. Lastly, Fig. (c) depicts another limitation of the geometric method. In this experiment, the direction of the pick-and-place task between the cutting board and the plate is predicted erroneously, which is not the case for the CNN-based method. In this case, the cutting board is moved under the plate, and since the cutting board is the object moved a bigger distance, it is detected as the picked object. Overall, the CNN-based method is considered a more comprehensive and accurate scene understanding method in analyzing pick-and-place tasks using initial and final images of a scene.

\section{Conclusion}
In this paper scene understanding in pick-and-place task detection using the initial and final shots from the scene was studied. An image dataset was collected and annotated for object detection and scene understanding tasks. Two methods for pick-and-place task detection were proposed and compared. The first method was based on geometric analysis of the bounding boxes of the objects in the scene, and the other was based on a CNN that could detect the spatial information between bounding boxes with overlap for both initial and final scenes separately and then make decisions accordingly. VGG16 proved to be the best feature extractor for spatial relationship information, outperforming ResNet-50 and ResNet-101. The experiments over unseen cases resulted in an accuracy of 84.3\% for the CNN-based method and 72\% for the geometric method. These experiments proved the advantage of the CNN-based method in detecting the pick-and-place tasks and their direction by about 12\%. In future work, the authors will employ the CNN-based method for a wider range of tasks, including pouring, cutting, stirring, etc. Also, more objects will be considered in the dataset to comprehensively include the possible tasks that a robot can perform. Moreover, scene understanding using continuous sequential data will be studied to address the rearrangement challenge using a video instead of two frames of a scene with spatial limits, and a mobile robotic manipulator.

% \bibliographystyle{IEEEtran}
% \bibliography{cas-refs.bib}

\begin{thebibliography}{99}
\bibitem{lenz2015deep}
I. Lenz, H. Lee, and A. Saxena, 
"Deep learning for detecting robotic grasps," 
\emph{The International Journal of Robotics Research}, vol. 34, no. 4-5, pp. 705–724, 2015.

\bibitem{hamed_primitive}
H. Hosseini, M. Koosheshi, M. T. Masouleh, and A. Kalhor, 
"Multi-Modal Robust Geometry Primitive Shape Scene Abstraction for Grasp Detection," 
submitted to \emph{Autonomous Robots, Springer}, 2023.

\bibitem{navidhand2023}
P. Yarmohammadi, N. Asadi Khomami, and M. T. Masouleh, 
"Experimental Study on Chess Board Setup Using Delta Parallel Robot Based on Deep Learning," 
submitted to \emph{International Conference on Robotics and Mechatronics}, 2023.

\bibitem{tamizi2022experimental}
M. G. Tamizi, A. A. A. Kashani, F. A. Azad, A. Kalhor, and M. T. Masouleh, 
"Experimental study on a novel simultaneous control and identification of a 3-DOF delta robot using model reference adaptive control," 
\emph{European Journal of Control}, vol. 67, p. 100715, 2022.

\bibitem{redmon2016you}
J. Redmon, S. Divvala, R. Girshick, and A. Farhadi, 
"You only look once: Unified, real-time object detection," 
in \emph{Proc. IEEE Conf. Computer Vision and Pattern Recognition}, pp. 779–788, 2016.

\bibitem{Roboflow}
B. Dwyer, J. Nelson, and J. Solawetz, 
"Roboflow (Version 1.0) [Software]," Available from \url{https://roboflow.com}, 2022.

\bibitem{deng2009imagenet}
J. Deng, W. Dong, R. Socher, L.-J. Li, K. Li, and L. Fei-Fei, 
"Imagenet: A large-scale hierarchical image database," 
in \emph{Proc. IEEE Conf. Computer Vision and Pattern Recognition}, pp. 248–255, 2009.

\bibitem{he2016deep}
K. He, X. Zhang, S. Ren, and J. Sun, 
"Deep residual learning for image recognition," 
in \emph{Proc. IEEE Conf. Computer Vision and Pattern Recognition}, pp. 770–778, 2016.

\bibitem{simonyan2014very}
K. Simonyan and A. Zisserman, 
"Very deep convolutional networks for large-scale image recognition," 
\emph{arXiv preprint arXiv:1409.1556}, 2014.

\bibitem{ai2thor}
L. Weihs, M. Deitke, A. Kembhavi, and R. Mottaghi, 
"Visual Room Rearrangement," 
in \emph{IEEE/CVF Conf. Computer Vision and Pattern Recognition (CVPR)}, June 2021.

\bibitem{zhou2014learning}
B. Zhou, A. Lapedriza, J. Xiao, A. Torralba, and A. Oliva, 
"Learning deep features for scene recognition using places database," 
\emph{Advances in neural information processing systems}, vol. 27, 2014.

\bibitem{cordts2016cityscapes}
M. Cordts et al., 
"The cityscapes dataset for semantic urban scene understanding," 
in \emph{Proc. IEEE Conf. Computer Vision and Pattern Recognition}, pp. 3213–3223, 2016.

\bibitem{song2022human}
Y. Song et al., 
"Human-in-the-loop Robotic Grasping using BERT Scene Representation," 
\emph{arXiv preprint arXiv:2209.14026}, 2022.

\bibitem{nguyen2019scene}
A. Nguyen, 
"Scene understanding for autonomous manipulation with deep learning," 
\emph{arXiv preprint arXiv:1903.09761}, 2019.

\bibitem{zech2017computational}
P. Zech et al., 
"Computational models of affordance in robotics: a taxonomy and systematic classification," 
\emph{Adaptive Behavior}, vol. 25, no. 5, pp. 235–271, 2017.

\bibitem{donahue2015long}
J. Donahue et al., 
"Long-term recurrent convolutional networks for visual recognition and description," 
in \emph{Proc. IEEE Conf. Computer Vision and Pattern Recognition}, pp. 2625–2634, 2015.

\bibitem{venugopalan2015sequence}
S. Venugopalan et al., 
"Sequence to sequence-video to text," 
in \emph{Proc. IEEE Int. Conf. Computer Vision}, pp. 4534–4542, 2015.

\bibitem{bohg2009grasping}
J. Bohg and D. Kragic, 
"Grasping familiar objects using shape context," 
in \emph{Proc. Int. Conf. Advanced Robotics}, pp. 1–6, 2009.

\bibitem{terven2304comprehensive}
J. Terven and D. Cordova-Esparza, 
"A comprehensive review of YOLO: From YOLOv1 to YOLOv8 and beyond," 
\emph{arXiv preprint arXiv:2304.00501}, 2023.

\bibitem{rahmania2022exploration}
R. Rahmania et al., 
"Exploration of The Impact of Kernel Size for YOLOv5-based Object Detection on Quadcopter," 
\emph{JOIV: Int. J. on Informatics Visualization}, vol. 6, no. 3, pp. 726–735, 2022.

\bibitem{zhao2022improved}
S. Zhao et al., 
"An improved YOLO algorithm for fast and accurate underwater object detection," 
\emph{Symmetry}, vol. 14, no. 8, p. 1669, 2022.

\bibitem{li2023yolo}
Z. Li et al., 
"A YOLO-GGCNN based grasping framework for mobile robots in unknown environments," 
\emph{Expert Systems with Applications}, vol. 225, p. 119993, 2023.

\bibitem{ye2023dynamic}
Y. Ye et al., 
"Dynamic and Real-Time Object Detection Based on Deep Learning for Home Service Robots," 
\emph{Sensors}, vol. 23, no. 23, p. 9482, 2023.

\bibitem{ahmad2020object}
T. Ahmad et al., 
"Object detection through modified YOLO neural network," 
\emph{Scientific Programming}, vol. 2020, pp. 1–10, 2020.

\bibitem{kshirsagar2023modified}
V. Kshirsagar, R. H. Bhalerao, and M. Chaturvedi, 
"Modified YOLO Module for Efficient Object Tracking in a Video," 
\emph{IEEE Latin America Transactions}, vol. 21, no. 3, pp. 389–398, 2023.

\bibitem{aljaafreh2023real}
A. Aljaafreh et al., 
"A Real-Time Olive Fruit Detection for Harvesting Robot Based on YOLO Algorithms," 
\emph{Acta Technologica Agriculturae}, vol. 26, no. 3, pp. 121–132, 2023.

\bibitem{hosseini2020improving}
H. Hosseini, M. T. Masouleh, and A. Kalhor, 
"Improving the successful robotic grasp detection using convolutional neural networks," 
in \emph{Proc. 2020 6th Iranian Conf. Signal Processing and Intelligent Systems (ICSPIS)}, pp. 1–6, 2020.

\bibitem{agile}
M. Koosheshi et al., 
"AGILE: Approach-based Grasp Inference Learned from Element Decomposition," 
accepted in \emph{Int. Conf. Robotics and Mechatronics}, 2023.

\bibitem{kolve2017ai2}
E. Kolve et al., 
"Ai2-thor: An interactive 3d environment for visual ai," 
\emph{arXiv preprint arXiv:1712.05474}, 2017.

\bibitem{wu2023prioritized}
Z. Wu et al., 
"Prioritized planning for target-oriented manipulation via hierarchical stacking relationship prediction," 
in \emph{Proc. 2023 IEEE/RSJ Int. Conf. Intelligent Robots and Systems (IROS)}, pp. 4873–4880, 2023.

\bibitem{zhu2018learning}
H. Zhu et al., 
"Learning tree-based deep model for recommender systems," 
in \emph{Proc. 24th ACM SIGKDD Int. Conf. Knowledge Discovery \& Data Mining}, pp. 1079–1088, 2018.

\bibitem{2024Naeinian}
F. Naeinian, E. Balazadeh, and M. T. Masouleh, 
"Mapping Human Grasping to 3-Finger Grippers: A Deep Learning Perspective," 
in \emph{Proc. 32nd Int. Conf. Electrical Engineering (in press)}, 2024.

\bibitem{balazade}
E. Balazadeh, M. T. Masouleh, and A. Kalhor, 
"HUGGA: Human-like Grasp Generation with Gripper’s Approach State Using Deep Learning," 
in \emph{Proc. 2023 11th RSI Int. Conf. Robotics and Mechatronics (ICRoM)}, pp. 854–860, 2023.

\bibitem{mojtahedi2024experimental}
M. Mojtahedi, A. Mohammadi, and M. T. Masouleh, 
"Experimental Study on Autonomous Food Packaging with Delta Parallel Robot and Two Fingered Gripper," 
in \emph{Proc. 32nd Int. Conf. Electrical Engineering}, 2024.

\bibitem{anushe}
A. Saadati, A. Kalhor, and M. T. Masouleh, 
"Deep Learning-Based Imitation of Human Actions for Autonomous Pick-and-Place Tasks," 
in \emph{Proc. 32nd Int. Conf. Electrical Engineering}, 2024.

\end{thebibliography}

\end{document}